\documentclass{article} 
\usepackage[authoryear,round]{natbib}
\usepackage[preprint]{colm}

\usepackage{microtype}
\usepackage{fontawesome5}
\usepackage{hyperref}
\usepackage{url}
\usepackage{booktabs}
\usepackage{svg}
\usepackage{amssymb} 
\usepackage{bbding}
\usepackage{tabularx}
\usepackage{tcolorbox}
\usepackage{csquotes}
\usepackage{epigraph}

\tcbuselibrary{listings,breakable}

\usepackage{tikz}
\usetikzlibrary{shapes.geometric, positioning, shadows}

\usepackage{lineno}
\usepackage{graphicx}
\usepackage{amsmath}
\usepackage{enumitem}
\usepackage{geometry}
\usepackage{xcolor}
\usepackage{xspace}
\usepackage{graphicx}
\usepackage{array} 
\usepackage{subcaption}
\usepackage{multirow}
\usepackage{xcolor}
\usepackage{colortbl}

\newif\ifshowvd
\showvdtrue    

\usepackage[colorinlistoftodos, textsize=scriptsize]{todonotes}

\definecolor{darkblue}{rgb}{0, 0, 0.5}
\definecolor{lightblue}{RGB}{173,216,230}
\definecolor{customlightblue}{HTML}{007cfa}
\hypersetup{colorlinks=true, citecolor=darkblue, linkcolor=darkblue, urlcolor=darkblue}
\usepackage{soul} 

\title{%
\centering
\makebox[\textwidth][c]{%
  \includegraphics[
      width=.35\textwidth,
  ]{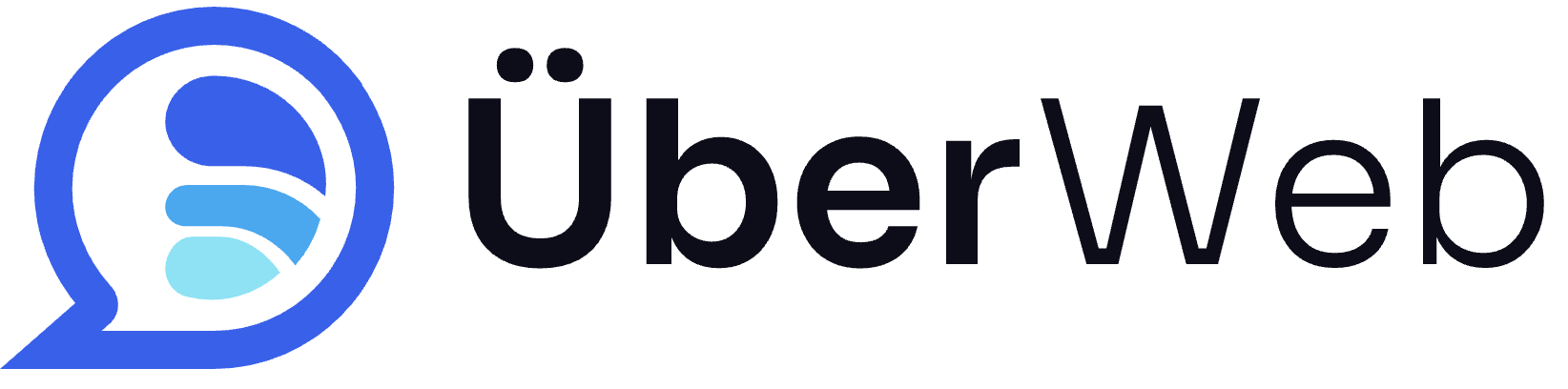}
}
Insights from Multilingual Curation\\
for a 20-Trillion-Token Dataset
}

\definecolor{LightCyan}{rgb}{0.88,1,1}

\usepackage{tabularx}   
\usepackage{array}      
\usepackage{ragged2e}   
\usepackage{svg}

\newcolumntype{L}{>{\RaggedRight\arraybackslash}X}

\begin{document}

\ifcolmsubmission
\linenumbers
\fi

\maketitle
{\vspace{-5.75em}
\begin{center}
\textbf{DatologyAI Team}\footnotemark
\end{center}
\footnotetext{See Contributions and Acknowledgments (\S~\ref{sec:contri}) for full author list.}

\begin{abstract}

Multilinguality is a core capability for modern foundation models, yet training high-quality multilingual models remains challenging due to uneven data availability across languages. A further challenge is the performance interference that can arise from joint multilingual training, commonly referred to as the “curse of multilinguality”. We study multilingual data curation across thirteen languages spanning multiple scripts, language families, and resource levels, showing that many reported regressions are not inherent to multilingual scaling but instead stem from correctable deficiencies in data quality and composition rather than fundamental capacity limits. In controlled bilingual experiments, improving data quality for any single language benefits others: curating English improves non-English performance on MMLU, ARC-Challenge, and Belebele in 12 of 13 languages (3.9\% average relative gain), while curating non-English yields reciprocal improvements in English (1.2\% average gain). Bespoke per-language curation produces substantially larger within-language improvements, with up to 16.9\% relative gains over uncurated baselines. Extending these findings to large-scale general-purpose training mixtures, we show that curated multilingual allocations comprising under 8\% of total tokens remain remarkably effective. We operationalize this approach within a broader large-scale effort that produced a 20T-token pretraining corpus derived entirely from public sources. Models with 3B and 8B parameters trained on a 1T-token random subset achieve competitive multilingual accuracy with 4–10× fewer training FLOPs than strong public baselines, establishing a new Pareto frontier in multilingual performance versus compute (Figure~\ref{fig:scaling_results}). Moreover, these benefits extend to frontier model scale: the 20T-token corpus served as part of the pretraining dataset for Trinity Large (400B/A13B), which exhibits strong multilingual performance relative to its training FLOPs. Together, these results show that targeted, per-language data curation mitigates multilingual interference and enables compute-efficient multilingual scaling.

\end{abstract}
\setlength{\epigraphwidth}{0.53\textwidth}

\epigraph{The future is already here – it's just not evenly distributed. }{--- William Gibson}

\section{Introduction}
\label{sec:intro}

Large language models (LLMs) have fundamentally reshaped the landscape of artificial intelligence, yet their benefits remain unevenly distributed across languages. Although modern models have demonstrated remarkable capabilities in English, these capabilities often degrade substantially when applied to non-English settings
\citep{ahuja2024megaverse, khanna2025invisible}. Bridging this gap is not merely an architectural challenge,
but fundamentally a data-centric one: training on large volumes of high-quality data is essential for achieving frontier-level model capabilities.
English benefits from multiple large-scale, carefully curated public corpora
\citep{penedo2024fineweb, li2024datacomp, su2025nemotron, olmo2025olmo},
whereas multilingual corpora are far more fragmented.
Many non-English languages occupy a long tail characterized by limited, noisy, or inconsistently curated data, constraining multilingual model performance regardless of architectural capacity.

\begin{figure}[!t]
    \centering
    \includegraphics[width=\textwidth]{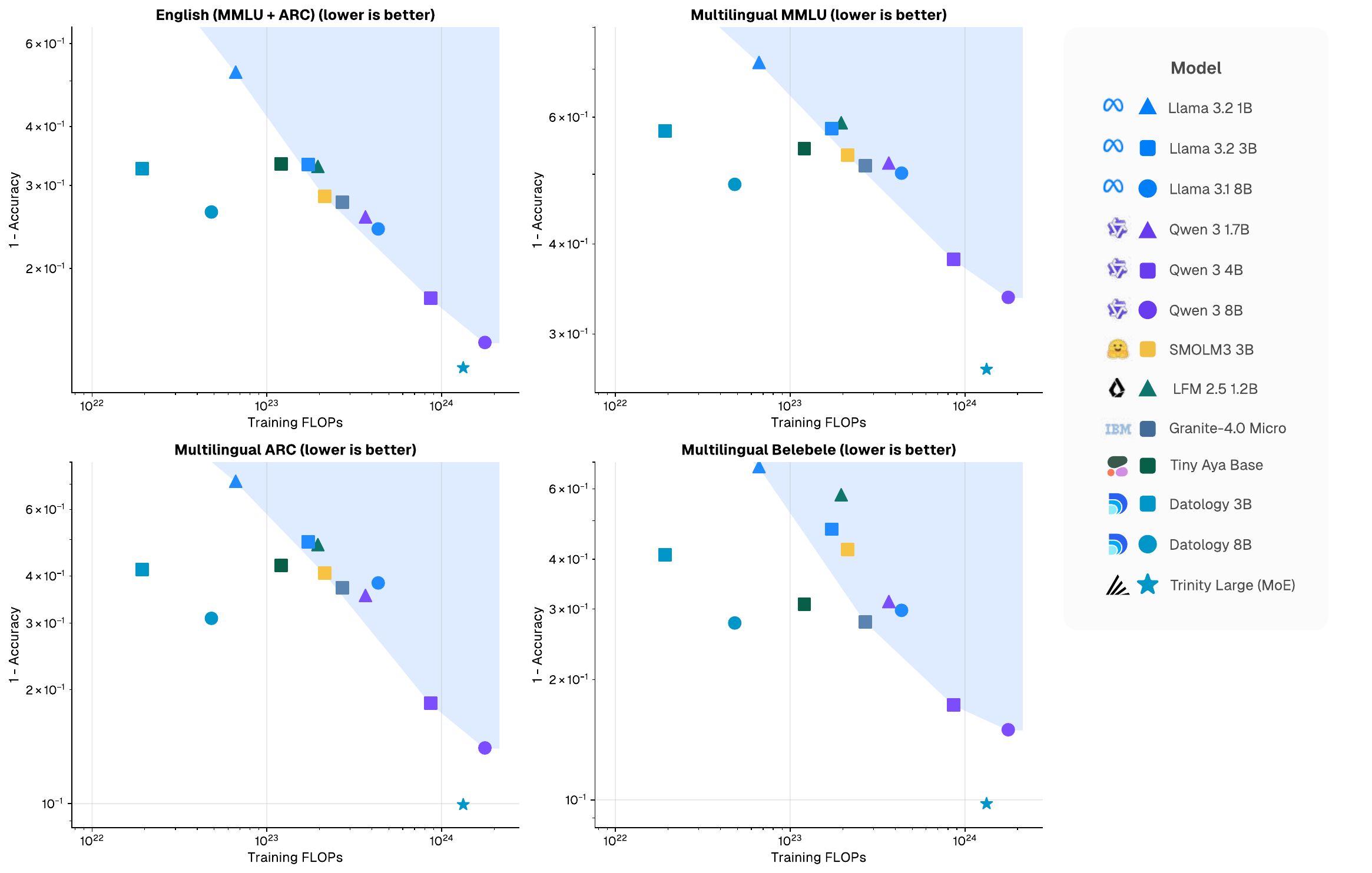} 
    \caption{\textbf{A new compute-performance Pareto frontier for English and multilingual capabilities}. We report 
    error rate (log-scale; 1-accuracy, lower is better) 
    as a function of training FLOPs (log-scale) across English (MMLU+ARC) and three multilingual
    benchmarks (Multilingual MMLU, Multilingual ARC, and Belebele).
    All evaluations use a multiple-choice format;
    multilingual scores are averaged over 13 languages.
    The shaded gray region summarizes the performance–compute envelope of representative open-weight baselines (e.g., Qwen3-4B/8B, Granite-4.0-3B).     
    DatologyAI models occupy the bottom-left region relative to these baselines, indicating substantially lower multilingual error at reduced compute. We restrict our English-language evaluations to MMLU and ARC-{Challenge} for parity with the multilingual evaluations, and reserve comprehensive English and quantitative benchmarking for forthcoming companion releases.
    }
    \label{fig:scaling_results}
\end{figure}


Beyond uneven data availability across languages, multilingual modeling faces an additional distinct challenge: the so-called ``curse of multilinguality" \citep{conneau2020unsupervised, chang2024multilinguality}. This term refers to the empirical observation that training a single model across an increasing number of languages often leads to degraded per-language performance, even under comparable training budgets.
Historically, the consensus view attributed this phenomenon to a capacity bottleneck, framing multilingual modeling as a zero-sum game in which distinct languages compete for finite parameters 
or model capacity
\citep{xue2021mt5,  conneau2020unsupervised, chang2024multilinguality}. Under this paradigm, the primary solutions have been to scale model size 
\citep{blevins2024breaking, pfeiffer2022xmod}
or increase the number of training tokens \citep{longpre2025atlas}, both of which substantially increase the computational cost of multilingual training. Such strategies, however, assume access to abundant, high-quality multilingual text, bringing them into direct tension with the uneven data availability across languages discussed above.

Recent evidence suggests this capacity-centric view is incomplete. Emerging research indicates that the ``curse" may stem less from parameter scarcity 
and more from the interference caused by suboptimal data quality. Both \cite{seto2025assessing} and \citet{revisiting2025mixtures} demonstrate that the trade-off between English and multilingual performance is not inevitable; they find that replacing significant portions of English data with high-quality multilingual text need not degrade English capabilities. Similarly, advances in model-based data selection \citep{messmer2025enhancing} and systematic filtering pipelines \citep{penedo2025fineweb2} reveal that when data quality is rigorously controlled, 
models can accommodate significantly more linguistic diversity without quality degradation. 
Taken together, these findings suggest that the apparent capacity constraints 
in multilingual scaling are often
induced by low-quality data. 
This motivates a shift in emphasis: 
optimal scaling of multilingual capabilities requires intentional, 
multilingual-targeted data curation. 



In this work, \textbf{we study multilingual foundation model training through the lens of data curation}, arguing that careful curation can simultaneously address the two central challenges of multilingual modeling: limited high-quality data for many languages and performance interference arising from joint multilingual training. By improving data quality, curation enhances cross-lingual transfer, reducing the amount of language-specific data required to achieve strong performance. The complement to this phenomenon is that targeted multilingual curation mitigates interference effects, alleviating the curse of multilinguality without relying solely on increasing compute.
We validate these claims through controlled 60B-token bilingual studies, large-scale 1T-token pretraining, and frontier-scale pretraining at multi-tens-of-trillions of tokens.

We summarize the key contributions of this work as follows:
\begin{enumerate}
    \item \textbf{Cross-lingual transfer improves with data quality.} We demonstrate that refining data quality drives significant cross-lingual performance gains
    through controlled bilingual experiments with 3B-parameter models trained on 60B tokens.
    Crucially, we find this relationship is bidirectional: 
    enhancing the quality of English data improves non-English performance in 12 out of 13 examined languages, yielding an average relative 
    improvement of 3.91\% across
    multilingual evaluations, while improving the quality of non-English data benefits English capabilities in 12 out of 13 languages, with an average relative improvement of 1.21\% on English evaluations.
    \item \textbf{Optimal performance requires bespoke multilingual curation.} 
    While English data curation does improve multilingual capabilities, we 
    find that the best performance is obtained when tailored curation 
    pipelines are built for each language. 
    Our findings highlight that  
    English-centric curation strategies cannot be applied blindly to other languages. Instead, it is imperative to construct tailored pipelines designed for each language's specific needs.
    For our 3B-parameter models trained on 60B tokens, while English curation alone drove the aforementioned 3.91\% relative improvement, applying bespoke language-specific curation yielded a significantly higher 16.87\% relative 
    improvement over the uncurated baseline
    \item \textbf{Data quality persists through translation.}
Building on findings of recent large-scale translation efforts \citep{wang2025mtpretraining, penedo2026finetranslations},
    we explore various strategies to translate English data into non-English languages.
    Large-scale translation provides a mechanism for expanding training data across languages, but we find that the choice of source data critically determines its effectiveness.
    We observe that prioritizing high-quality English documents for translation can significantly boost performance over 
    translations of arbitrary English documents.
    In experiments with 3B-parameter models, we find that augmenting the uncurated baseline with translations of random English data yields marginal gains, whereas translating high-quality, score-filtered English data leads to an average relative 
    improvement of 5.09\% over an uncurated baseline.
    Moreover, we find that translation is most effective when embedded within a holistic, per-language curation framework, which yields the strongest overall performance.
    \item \textbf{Curation makes multilingual scaling remarkably compute-efficient.}
    Under a 1T-token training budget drawn from a curated general-purpose pretraining corpus, we find that allocating approximately 8\% of tokens to high-quality multilingual data ($\sim$80B tokens across 13 languages) is sufficient to achieve very strong multilingual performance, in many cases comparable to or exceeding competitive open-weight models.
    Our 3B and 8B models trained for 1T tokens achieve 4–10$\times$ greater training FLOPs efficiency than strong public baselines. For example, a DatologyAI 3B model trained for 1T tokens ($1.8 \times 10^{22}$ FLOPs) outperforms LFM-2.5-1.2B \citep{LiquidAI2026}, a 1.2B model trained for 28T tokens ($1.9 \times 10^{23}$ FLOPs). Similarly, a DatologyAI 8B model trained for 1T tokens ($4.8 \times 10^{22}$ FLOPs) outperforms SmolLM3-3B \citep{bakouch2025smollm3} and Granite-4.0-3B \citep{ibm_granite3_0_techreport}, both trained with an order of magnitude more compute. Importantly, \textbf{these efficiency gains persist at frontier scale:} our multilingual curation framework forms part of the 17T-token pretraining corpus for Trinity Large Base (400B-parameter MoE, 13B active; \citet{TrinityLarge}), which exhibits exceptionally strong multilingual performance for its training FLOPs budget.
    
    
\end{enumerate}


Our results demonstrate the critical role of multilingual data curation for multilingual model capability. From controlled 60B-token studies to 1T-token training mixtures to 17T-token frontier-scale pretraining, we show that language-aware improvements in data quality systematically enhance cross-lingual transfer, mitigate multilingual interference, and substantially improve within-language performance. These effects collectively shift the performance–compute Pareto frontier (see Figure~\ref{fig:scaling_results}). Together, these findings position high-quality, per-language curation as a practical and scalable mechanism for compute-efficient multilingual foundation model training, advancing progress toward more language-inclusive foundation models and a more evenly-distributed future.

\section{Related Work}

\paragraph{Multilingual Data Curation.}
The field has moved from scale-focused web ingestion \citep{xue2021mt5,suarez2020oscar} to systematic, reproducible data curation that prioritizes quality, auditing, and strong filtering to reduce noise and contamination.
Recent efforts such as FineWeb \citep{penedo2024fineweb} and FineWeb2 \citep{penedo2025fineweb2} have formalized the curation process, releasing reproducible pipelines that scale high-quality filtering across thousands of languages. 
In addition  to these advances in pre-training, the Aya initiative \citep{singh2024aya} presents a multilingual
post-training dataset focused on instruction-following across 65 languages.

Beyond the construction of large-scale multilingual corpora, there have also been recent efforts to 
improve the quality of these corpora via multilingual curation; 
both \cite{messmer2025enhancing} and \cite{chen2025muratinghighqualitydata}
propose general purpose model-based filtering solutions to improve multilingual data quality. 
There are also examples of highly specialized, language-specific curation efforts such as 
\cite{burns2025alephalphagermanwebimprovinggermanlanguagellm} for German 
and \cite{khan2024indicllmsuite} for Indic languages.
Our focus in this work is closely aligned with these multilingual curation efforts, and helps to further
emphasize the performance improvements which can be unlocked via data curation.


\paragraph{Data Mixing and Interference.}
While the existence of cross-lingual transfer is well established \citep{pires2019multilingual}, a central challenge remains how to drive positive transfer across
languages while minimizing negative interference
\citep{conneau2020unsupervised, wang2020neginterf}. 
While temperature-based sampling is a standard heuristic \citep{conneau2019xlm}, it often leads to overfitting in low-resource regimes. Strategies like UniMax \citep{chung2023unimax} address this by capping repetition to ensure more representative coverage. 
A further area of research is 
the use of dynamic curricula
to drive 
multilingual performance; 
\cite{choi2023order} advocate for a two-stage training paradigm which 
first pre-trains on high resource languages and subsequently 
fine-tunes on lower resource languages. Conversely, 
 \cite{revisiting2025mixtures}
arrive at the conclusion
that staging the introduction of languages does not
yield tangible improvements.
While this work contains some examination of curricula and multilingual mixture proportions, the 
central theme is demonstrating that careful 
curation can significantly improve cross-lingual transfer dynamics, thus reducing interference.

\paragraph{Multilingual Scaling Laws.}
While scaling behaviors for English-centric models are well-characterized \citep{hestness2017deep, kaplan2020scaling,hoffmann2022training}, extending these laws to the multilingual setting introduces significant complexity due to cross-lingual transfer dynamics. 
Early attempts primarily focused on machine translation \citep{fernandes2023scaling}, but recent work has targeted general-purpose decoder-only architectures. \citet{he2025scaling} propose a ``family-based" scaling law, demonstrating that the test loss for a language family is primarily determined by its own sampling ratio, largely independent of other families in the mixture. This simplifies the analysis of inter-language competition but does not fully account for the ``curse of multilinguality" phenomena observed when scaling to many languages. 
Addressing this, the ATLAS project recently conducted the largest study to date, covering over 400 languages and exploring cross-lingual transfer across 38 languages \citep{longpre2025atlas}.
Their work derives an adaptive transfer scaling law that explicitly models the trade-off between adding languages and maintaining performance per parameter. This provides a first-principles guide for optimal capacity allocation in massively multilingual settings.
While these laws focus on parameter-based trade-offs, 
our results instead demonstrate that careful, 
language-specific curation allows 
us to significantly improve on current scaling laws by shifting the 
bottleneck from model capacity to data quality. 
In this way, we are able to demonstrate that the 
``curse of multilinguality" 
is the result of correctable 
deficiencies in the training data.


\section{Experimental Setup and Methodology}

\noindent\textbf{Pretraining Data.} In this work we curate exclusively on top of open source corpora. For English, we leverage the DCLM corpus \citep{li2024datacomp}, FineWeb \citep{penedo2024fineweb}, and the non-synthetic components of Nemotron CC v1 \citep{su2025nemotron}. For non-English data, we rely on the FineWeb2 corpus \citep{penedo2025fineweb2}. While FineWeb2 supports over 1{,}000 languages, in this work we focus on a set of 13 diverse non-English languages spanning multiple writing systems and language families (see Table \ref{tab:languages}).


\begin{table}[t]
\centering
\footnotesize
\setlength{\tabcolsep}{4pt}
\begin{tabular}{lllrr}
\toprule
\textbf{Language} & \textbf{Family} & \textbf{Script} & \textbf{FineWeb2 Documents (M)} & \textbf{Llama 3.2 1B Tokens (B)} \\
\midrule
Russian    & Slavic        & Cyrillic           & 699.1 & 1004.6 \\
Chinese    & Sino-Tibetan  & Hanzi        & 636.1 & 743.4 \\
German     & Germanic      & Latin              & 496.0 & 407.0 \\
Spanish    & Romance       & Latin              & 441.3 & 352.3 \\
Japanese   & Japonic       & Kanji + Kana & 400.1 & 404.4 \\
French     & Romance       & Latin              & 360.1 & 306.4 \\
Portuguese & Romance       & Latin              & 199.7 & 160.1 \\
Indonesian & Austronesian  & Latin              & 100.2 & 101.8 \\
Arabic     & Semitic       & Arabic             & 62.0 & 63.5 \\
Vietnamese & Austroasiatic & Latin              & 61.1 & 47.8 \\
Korean     & Koreanic      & Hangul             & 60.9 & 59.5 \\
Hindi      & Indo-Aryan    & Devanagari         & 22.1 & 25.1 \\
Bengali    & Indo-Aryan    & Bengali            & 15.2 & 38.7 \\
\bottomrule
\end{tabular}
\caption{\textbf{Non-English languages included in this study.}}
\label{tab:languages}
\end{table}

The languages above also span a wide range of resource levels in publicly available web text: Spanish is high-resource (with hundreds of billions of available tokens), whereas Hindi, Bengali, and Arabic are comparatively low-resource, making them particularly sensitive to data scarcity and quality. 

\noindent\textbf{Data curation.} Building on our work at DatologyAI, we develop language-specific data curation pipelines for each of the languages above. For English, we build on our state-of-the-art web curation pipeline \citep{datologyai_text_2024}, which integrates complementary strategies including model-based filtering, embedding-based selection, and targeted synthetic data generation \citep{beyondweb}.
For each non-English language, we tailor our curation pipeline to language's linguistic and distributional characteristics rather than directly applying the English recipe. Concretely, this includes selecting, validating, and/or training language-appropriate models for 1) filtering, 2) embedding for geometry-based curation, and 3) synthetic rephrasing. We also adapt filtering and mixing strategies to account for script- and language-specific artifacts and varying token scarcity across languages.
To quantify the impact of these interventions, we compare to \textbf{uncurated baselines}, defined throughout this work as samples drawn at random from
the DCLM for uncurated English and  FineWeb2 for uncurated non-English corpora\footnote{We note that DCLM and FineWeb2 were heavily curated as part of their development, and use the term ``uncurated" as meaning ``not subject to DatologyAI curation".}.

\noindent\textbf{Model.} We present results on both 3B and 8B parameter models using a Llama-based architecture \citep{touvron2023llama}. 
Throughout this work we use the Llama-3.2 tokenizer. 
All models are trained and evaluated with a context window of 4096 tokens.
We note that because the focus of this work is on the effects of data curation, we did not attempt to optimize model quality via any means other than data curation. All models of a given size used identical training configurations in every way except the dataset.

\noindent\textbf{Evaluation.} We evaluate our models on three complementary multilingual benchmarks:
\begin{itemize}
    \item \textbf{Multilingual MMLU \citep{singh2025global}:} measures broad knowledge and academic-style reasoning across diverse subject areas, including STEM, humanities, and social sciences.
    \item \textbf{Multilingual ARC Challenge \citep{lai2023okapi}:} measures multi-step reasoning on grade-school science questions; it covers a narrower domain than MMLU but places greater emphasis on compositional reasoning. 
    \item \textbf{Belebele \citep{bandarkar2024belebele}:} measures multilingual reading comprehension and semantic reasoning over aligned passages, with minimal dependence on memorized factual knowledge.
\end{itemize}
We complement these multilingual evaluations with English MMLU and ARC evaluations.\footnote{We restrict our English-language evaluations to MMLU and ARC-{Challenge} for parity with the multilingual evaluations, and reserve comprehensive English and quantitative benchmarking for forthcoming companion releases.}  
Throughout this work, we
rely on the \textbf{lighteval} \citep{lighteval} framework for all our evaluations. We report zero-shot performance. For large-scale experiments (e.g., Figure~\ref{fig:scaling_results}), we adopt the multiple-choice formulation (MCF), following common practice \citep{gu2025olmes,li2024datacomp}. For smaller runs (e.g., our 3B, 60B-token setting), we instead use the cloze formulation. This choice is motivated by statistical efficiency: cloze-style scoring yields a denser learning signal and typically reduces variance relative to discrete option selection, making it better suited for low-resource or early-training regimes where multiple-choice accuracy can be dominated by near-random guessing \citep{gu2025olmes,li2024datacomp}. 
Finally, we note that all models in this manuscript are base (pretrained) models and were evaluated without any post-training or fine-tuning.
A full list of evaluation datasets by language is provided in Appendix~\ref{app:evals_per_lang}.

\section{Main Findings}
\label{sec:main-results}

\subsection{The impact of curation on multilingual transfer dynamics and language-specific performance}
\label{sec:bilingual-exps-part1}

\emph{Cross-lingual transfer} 
refers to the observation that improving representations in one language can benefit performance in other languages. As models scale and language coverage increases, 
such transfer becomes increasingly important.
In this section, we investigate the impact of data quality on cross-lingual transfer and identify opportunities to improve downstream performance via data 
curation 
interventions on both English and non-English data. 

\subsubsection{Improving English data quality improves cross-lingual performance}
\label{bilingual-section1}

Most multilingual language models are trained on predominantly English corpora, making English data quality a central determinant of multilingual performance. Yet the extent to which English data quality governs cross-lingual transfer remains insufficiently characterized. In a series of controlled experiments, we show that English-to–non-English transfer is strongly mediated by the quality of the English training data.
In particular, 
improving the quality of the English portion of training data mixtures 
yields consistent performance gains across 
almost all non-English languages considered.
We train a suite of 3B-parameter models for 60B tokens under a range of dataset compositions, focusing on bilingual settings consisting of English paired with a single ``target" language. This design yields 13 language pairs (e.g., English–Spanish, English–German, etc), each trained with a fixed 50/50 mixture ratio. For every pair, we compare three curation regimes: 
\begin{enumerate}[label=\roman*.]
    \item Uncurated English DCLM and uncurated FineWeb2 non-English data (i.e., random samples from DCLM and FineWeb2).
    \item DatologyAI-curated English and uncurated FineWeb2 non-English data.
    \item DatologyAI-curated English and DatologyAI-curated non-English data.
\end{enumerate}

\noindent \textbf{Cross-lingual impact of English curation.}
Figure \ref{fig:curation_impact_bilingual} summarizes performance across 13 languages, reporting average scores over multilingual MMLU, ARC Challenge, and Belebele. English-only curation (light blue bars) yields consistent gains over the uncurated baseline (dark purple bars) in every language except Bengali, indicating that improving English data quality alone can measurably strengthen multilingual capabilities in otherwise uncurated languages. 
Averaged across languages, English curation yields a {3.91\%} relative improvement in non-English performance compared to an uncurated baseline.


\begin{figure*}[h!]
    \centering
    \includegraphics[width=\textwidth]{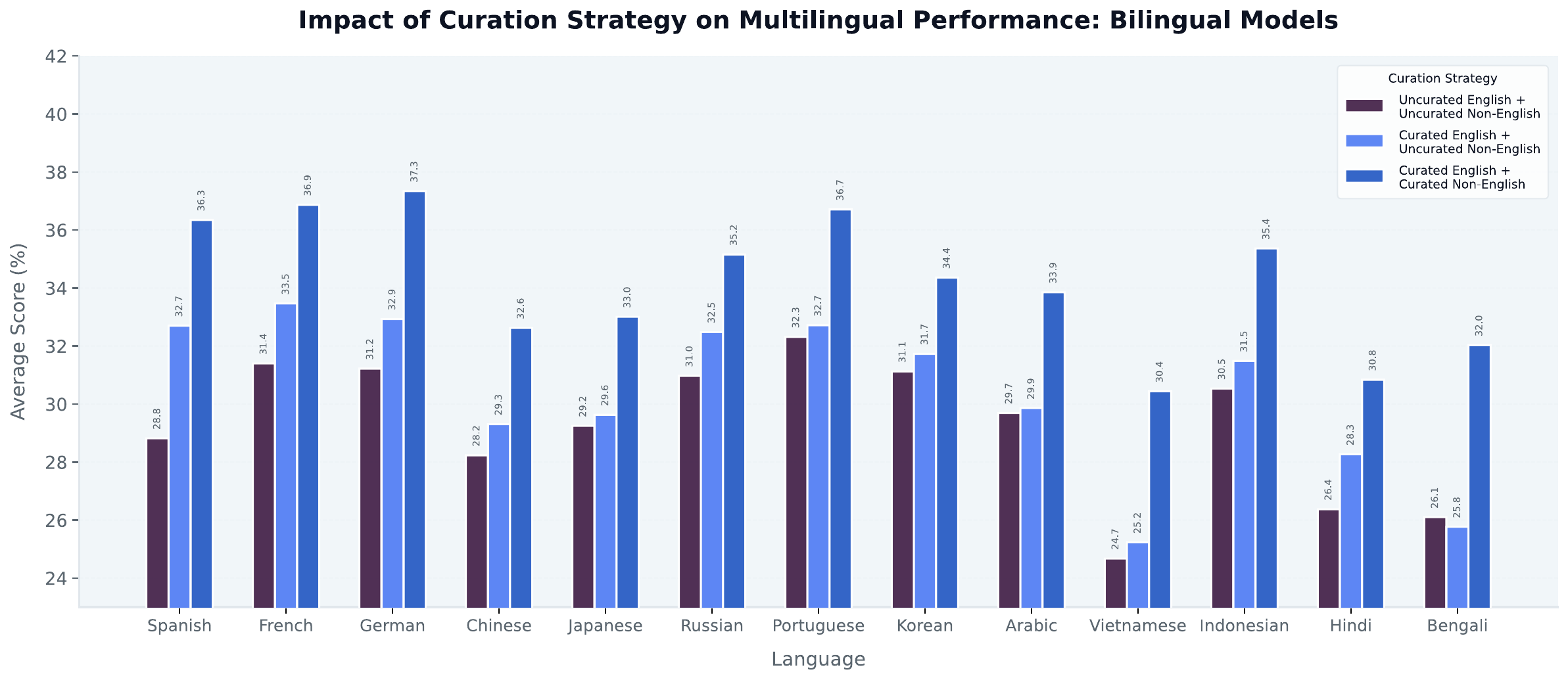}
    \caption{\textbf{Impact of Curation Strategy on Multilingual Performance (bilingual models).} 
    Performance comparison for 3B parameter models trained on 60BT tokens (50:50 English:non-English ratio). Results are averaged across multilingual MMLU, ARC, and Belebele.
    Across 13 languages, results show that improved English curation (light blue bars) consistently improves performance over the uncurated baseline (dark purple bars; improvement in 12 of 13 languages), while combining curated English with curated multilingual data (dark blue bars) yields the highest average scores across all languages.}
    \label{fig:curation_impact_bilingual}
\end{figure*}

\subsubsection{Cross-lingual curation gains correlate with language similarity}

While English curation improves performance in the uncurated language for 12 out of 13 examined languages, the magnitude of these benefits is not uniform. 
Languages such as Spanish, French, and German, which are linguistically more similar to English, exhibit more pronounced uplifts than
languages such as Hindi and Arabic (8.56\% compared to 3.94\% relative gains, respectively).
This finding is in line with \citet{longpre2025atlas}, who report that bilingual transfer (as measured by cross-entropy loss) is predicted by language similarity across varying sampling ratios.

We ask a similar, though distinct question here: what is the relationship between language similarity and the impact of English curation on non-English \emph{model capabilities}.
We consider two heuristic approaches to quantify linguistic distances: similarity
in embedding space and perplexity. Crucially, to ensure these metrics capture linguistic divergence rather than topical shifts in the underlying text, we compute both measures on parallel samples from the FLoRes dataset \citep{goyal2022flores}.
For embedding distance, we report the average cosine distance between English and the target language across three distinct models: LaBSE \citep{feng2022language}, e5-small \citep{wang2022text}, and sentence-transformers \citep{reimers2019sentence}.
Our perplexity proxy is defined as the average log perplexity per word as measured on the target language samples under a model trained exclusively on curated English data. We explicitly do not normalize by word length, allowing this metric to serve as a raw measure of how well English-centric patterns generalize to the target distribution.

Figure \ref{fig:similarity_transfer} illustrates 
a significant negative correlation between 
proxy distance metrics and the relative improvement gained through English curation alone. Specifically, embedding 
distance yields a Pearson correlation of $-0.62$ ($p=0.024$), 
while perplexity shows an even stronger correlation of $-0.70$ ($p=0.018$). 
These results 
provide evidence that language similarity, quantified using 
two distinct approaches, 
is significantly correlated with the 
cross-lingual gains from English-only curation.


\begin{figure}[!htbp]
    \centering
    \includegraphics[width=.98\textwidth]{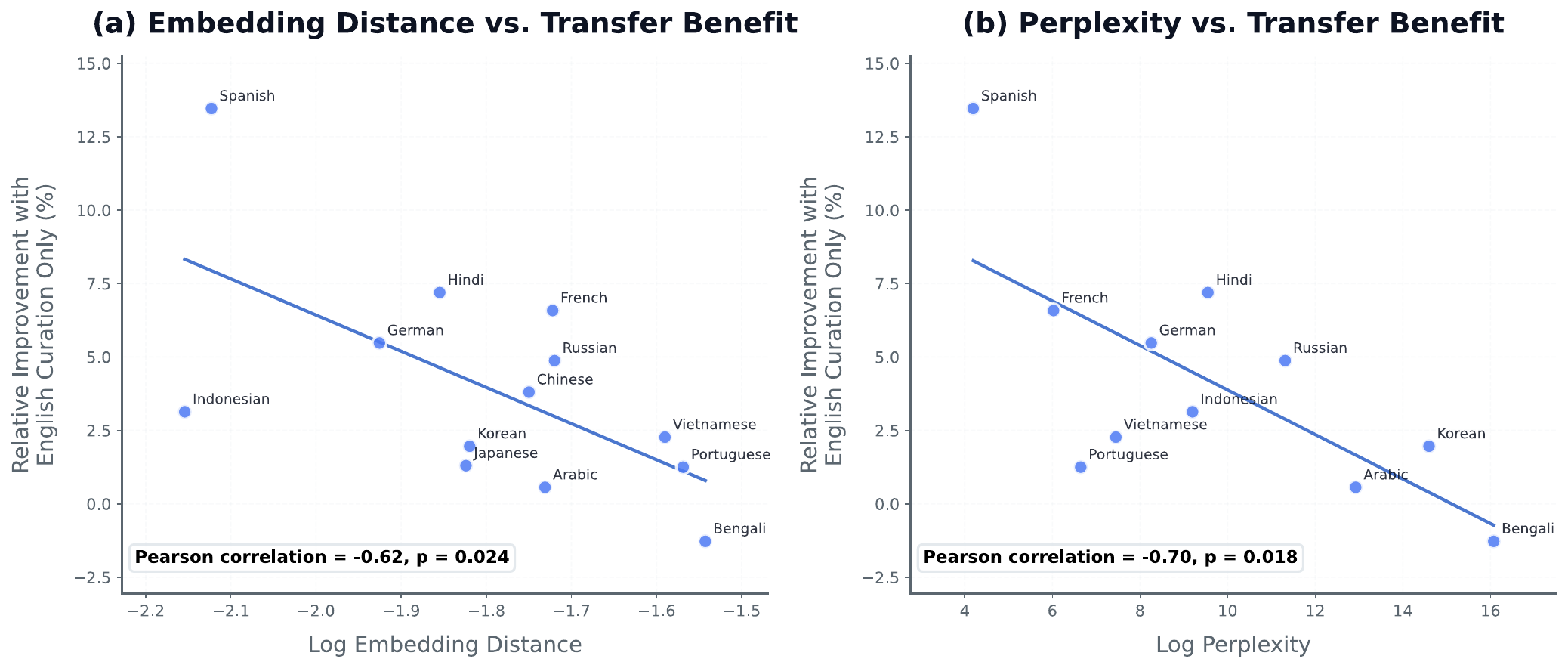}
    \caption{\textbf{Correlation between language similarity to English and cross-lingual transfer benefit.} We evaluate linguistic distance using two proxies: (a) average log embedding distance across LaBSE, e5-small, and sentence-transformers, and (b) log perplexity of the target language under an English-only model. Both metrics show a significant negative correlation (Pearson $r=-0.62$ and $r=-0.70$ respectively) with the performance uplift gained from English data curation. These results demonstrate that linguistically similar languages, such as Spanish and French, receive the most pronounced benefits from high-quality English data, while more distant languages like Bengali and Arabic show significantly lower transfer gains.
    }
    \label{fig:similarity_transfer}
\end{figure}

\subsubsection{Optimal multilingual performance requires bespoke multilingual curation}

While curating English consistently improves cross-lingual performance, it is not sufficient to reach optimal performance in any given target language. Figure~\ref{fig:curation_impact_bilingual} shows a persistent gap in performance when non-English data is curated (dark purple) versus when it is not (purple and light blue). 
Moreover, we note that the magnitude of improvements from curating for each language individually far exceeds the benefits associated with only curating English data. 
This finding, while not necessarily surprising, further emphasizes the need to pay careful attention to data quality and curation for each language individually. 
This observation reinforces the argument by \citet{messmer2025enhancing} that generic, English-centric heuristics will not generalize 
across diverse alphabets and scripts. 
These results are also consistent 
with  \citet{revisiting2025mixtures}, who posited that 
it is the presence of noisy, uncurated data which harms multilingual models rather than an issue of model capacity. That is, the issue 
``resembles a curse of \textit{data quality}" rather than a curse of multilinguality.

\subsubsection{Improved non-English data curation also benefits English capabilities}

Prior sections highlighted
that while English curation improves non-English performance, 
optimal non-English performance comes from careful curation of 
both English and non-English data. 
In this section we study the effect of multilingual data curation on English performance.
Despite many recent findings, we observe that the benefits of data curation 
are bidirectional: 
our results demonstrate that improving the quality of the \textit{non-English} data component also yields consistent gains on English benchmarks. Figure~\ref{fig:english_impact} compares the performance on English tasks (MMLU and ARC Average) between models trained with uncurated versus curated non-English data, while keeping the English data constant (we use curated English data throughout).
The results demonstrate an average relative improvement of 1.2\%. Concretely, in
12 out of the 13 language considered, 
the bilingual model trained with fully curated data (i.e., both English and non-English data curated) outperformed the
version with uncurated non-English data.

\begin{figure}[h!]
    \centering
    \includegraphics[width=\textwidth]{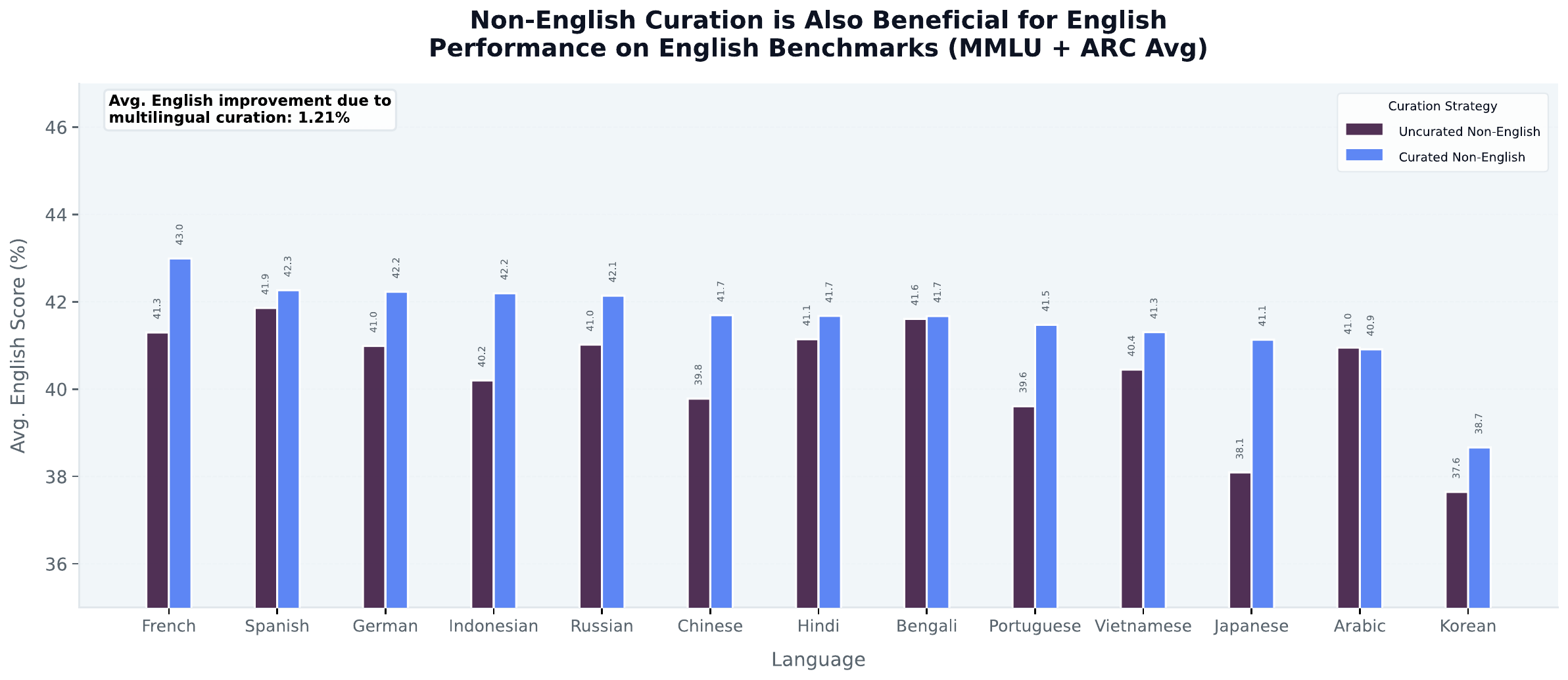}
    \caption{\textbf{Non-English Curation Benefits English Performance.} 
    Performance comparison for 3B parameter models trained on 60BT tokens (50:50 English:non-English ratio). 
    Results are average of English MMLU and ARC. We contrast performance when
    when the accompanying multilingual data is uncurated (dark purple) versus curated (dark blue).  We observe positive transfer in 12 out of 13 languages, with an overall relative improvement of 1.21\%.  
    }
    \label{fig:english_impact}
\end{figure}

At a high-level, these findings suggest that high-quality data acts as a globally beneficial signal in model training, providing a means to mitigate the ``curse of multilinguality'' by systematically improving the quality of data for each individual language.


\subsection{The efficacy of translation as augmentation is determined by source quality}
\label{sec:bilingual-exps-part2}

Machine translation has increasingly surfaced as a viable strategy for enhancing multilingual model performance, 
serving as a scalable source of synthetic data \citep{seto2025assessing, wang2025mtpretraining}. 
Prior efforts, such as FineTranslations \citep{penedo2026finetranslations}, have successfully utilized large-scale translation pipelines to map multilingual content into English with an explicit focus of improving 
translation capabilities. In this work, we instead investigate the 
effectiveness of translation as a general tool to drive overall
multilingual capabilities.
We demonstrate that this strategy can indeed drive  performance gains; however, its effectiveness is heavily contingent on source data quality. Our results indicate that translating only high-quality documents, selected via score filters, leads to markedly better improvements.

To understand how to best leverage translation 
as a tool within multilingual data curation, 
we conducted controlled experiments on three languages: Hindi, Bengali, and Arabic. 
We trained 3B parameter models for 60B tokens, keeping the English component fixed and curated, while varying the non-English strategy. 
We compared three approaches: (1) an uncurated baseline, (2) augmenting the baseline with translations of randomly selected English data (i.e., blind translations), and (3) augmenting with translations of high-quality, scored English data. When scoring English data, we used a fasttext classifier similar to \cite{penedo2024fineweb}. 

\begin{figure}[h!]
    \centering
    \includegraphics[width=\textwidth]{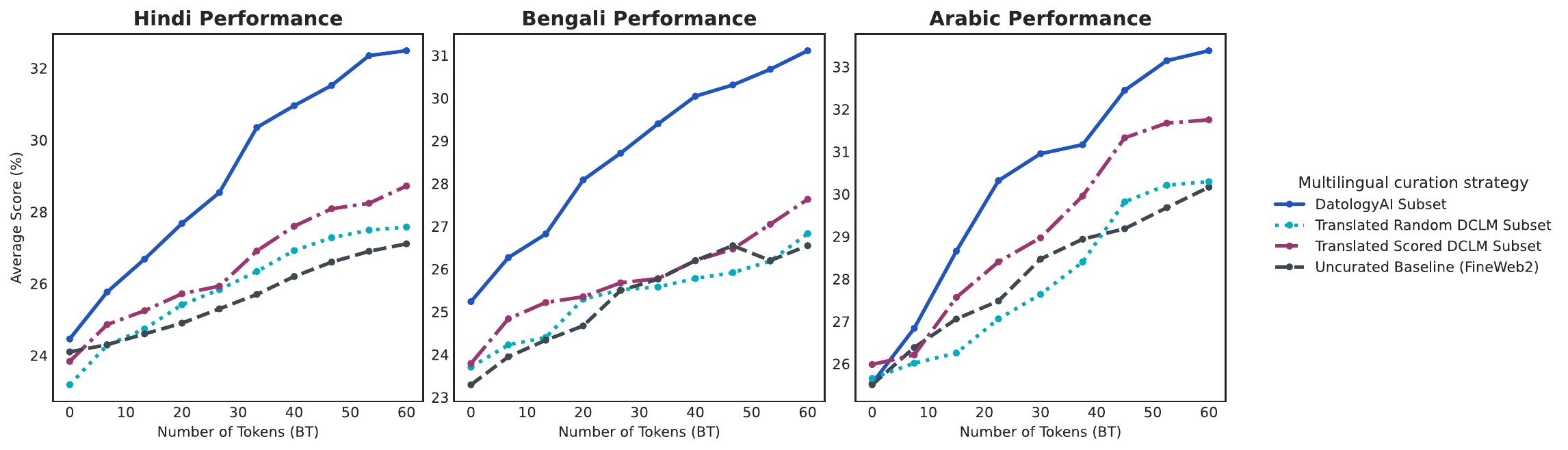}
    \caption{\textbf{Evaluation of benefits associated with Random vs Scored Translation for Low-Resource Languages.} Performance curves for Hindi, Bengali and Arabic showing that while augmenting training data with translated English text (red and cyan lines) improves over the uncurated baseline (dark gray), it still falls short of the performance achieved by bespoke DatologyAI curation (dark blue).}
    \label{fig:translation_performance}
\end{figure}

The results, illustrated in Figure~\ref{fig:translation_performance}, reveal a clear hierarchy of performance. Augmenting the uncurated baseline with 
arbitarily translated English data
only yields marginal performance improvements over a purely 
 uncurated baseline. However, the magnitude of improvements grows when translating high quality, score-filtered data. This 
result mirrors findings in \cite{beyondweb}, which showed that the quality of input documents for synthetic rephrasing is crucial to obtaining strong performance.

However, a significant performance gap remains. Our bespoke curation strategy (dark blue line) substantially outperforms both the uncurated baseline and the translation-augmented models. 
These findings imply that while translation 
can be a valuable component of multilingual data curation, as reported by \cite{wang2025mtpretraining}, its efficacy is ultimately determined by source data quality, and the best performance is obtained with a holistic curation approach across all target languages.

\subsection{Integrating multilingual curation into a general pretraining mix}
\label{section:big_boi_exps}

Sections \ref{sec:bilingual-exps-part1} and \ref{sec:bilingual-exps-part2} presented smaller-scale, controlled experiments 
intended to dissociate the impact of different multilingual
curation choices. 
However, an open question is how such 
curation strategies can scale
to larger token budgets and models, and how multilingual curation 
interacts with general purpose 
curation. To that end, 
we curated a 20T token dataset intended for foundation model training and frontier capabilities across English, multilingual, code, STEM, and reasoning skills.
The curation included generating over 8T tokens of synthetic English and non-English web data, and code and STEM data.
A random 1T subset of this
dataset was used to train both 3B and 8B parameter models following a Llama architecture.

\noindent\textbf{Multi-Phase Data Curriculum.}
To balance multiple diverse data streams, we implemented a multi-phase training curriculum that progressively increases the density of multilingual tokens. 
The mixture of tokens across three phases followed that used in the 
Trinity Large model \cite[Section 3.1]{TrinityLarge}.
The training process was divided into three distinct phases: 
\begin{itemize}
\item \textbf{Phase 1:} 650B tokens with 5\% multilingual data 
\item \textbf{Phase 2:} 250B tokens with 10\% multilingual data 
\item \textbf{Phase 3:} 100B tokens with 20\% multilingual data \end{itemize} 
Across the full training duration, this schedule resulted
in an overall allocation of 7.75\% tokens to our multilingual curation pipeline, supporting 
13 languages and thus resulting in an average of 6B tokens per langauge.
Despite this seemingly modest multilingual budget, our bespoke curation strategy allows 
models to obtain competitive performance across 
diverse languages spanning Latin, Cyrillic, Arabic, Indic, and CJK scripts. 



\paragraph{Establishing a New Pareto Frontier.}
\autoref{fig:scaling_results} positions DatologyAI models
against several
open-weights models across  
English (MMLU + ARC Average) and three multilingual benchmarks (Multilingual MMLU, ARC, and Belebele). 
The y-axis denotes the error rate ($1 - \text{Average Accuracy}$) in log scale, where lower values indicate superior capabilities. The shaded gray region encapsulates the performance-compute trade-off established by leading open-weights baselines, 
including 
Qwen3 \citep{yang2025qwen3}, Granite \citep{ibm_granite3_0_techreport}, SmolLM3 \citep{bakouch2025smollm3}, LFM-2.5 \citep{LiquidAI2026}, 
and Tiny Aya \citep{cohere2026tinyaya}. Our results demonstrate a marked shift in efficiency: the 
DatologyAI models 
consistently improve upon the established Pareto frontier. By
achieving error rates comparable to significantly larger and/or more compute-intensive baselines, we effectively redefine the Pareto frontier for multilingual foundational models.

\paragraph{Data curation unlocks token efficient data mixtures.}
We emphasize that the DatologyAI models in Figure~\ref{fig:scaling_results} use the smallest multilingual data mixture among all models that report mixture composition. Specifically, DatologyAI allocates only 7.75\% 
of training tokens to multilingual data while supporting a substantially 
broader set of languages. This proportion is markedly lower than those reported by comparable baselines, including LFM, which uses 20\% multilingual tokens \citep{LiquidAI2026}, and SmolLM3, which employs a 12\% multilingual mixture \citep{bakouch2025smollm3}.


In Appendix \ref{app:perf_per_language}, we report per-language performance breakdowns for each evaluation. We present
results across three groups of languages: 
Latin-script languages (Figure \ref{fig:per_language_latin}), Indic and Arabic (Figure \ref{fig:per_language_indic_arabic}), and Chinese, Japanese, Korean, and Russian (Figure \ref{fig:per_language_cjk}). Across all 13 languages, DatologyAI’s multilingual curation consistently improves the performance–compute Pareto frontier, yielding higher accuracy at a given FLOP budget (or comparable accuracy with less compute). 
We also present results comparing to various language-specialized base models, i.e. 
models
that have a focus on achieving strong performance on particular languages.
Examples include the  
Sarvam-1 model \citep{sarvamai_sarvam1_blog_2024}, focused on Indic languages, in Figure~\ref{fig:per_language_indic_arabic}; Trillion Labs Tri-7B \citep{trillionlabs_tri7bbase_2025}, focused on Korean, Japanese, and Chinese, in Figure~\ref{fig:per_language_cjk}; and SEA-LION-v3-9B 
\citep{ng2025sealion}, focused on Southeast Asian languages, also in Figure~\ref{fig:per_language_cjk}.
The specialized models also significantly improve upon the Pareto frontier, but their performance is comparable to that of the DatologyAI models on the particular languages they focus on; for example, Figures~\ref{fig:per_language_indic_arabic}
and Figures~\ref{fig:per_language_cjk}
show that DatologyAI models can meet or exceed performance 
of specialized models such as Sarvam-1 and Tri-7B, which are trained using similar or larger FLOPs budgets. 

The rightmost columns in Figures~\ref{fig:per_language_latin}--\ref{fig:per_language_cjk} illustrate the relationship between language-specific performance and aggregate multilingual proficiency. DatologyAI models consistently align with the line of unity, reflecting a data curation strategy
that prioritizes broad multilingual parity over individual language optimization. In contrast, specialized models like Sarvam-1 and Tri-7B exhibit a clear departure from this trend, appearing  above the line of unity for their target languages. However,
their aggregate multilingual performance (shown along x-axis) reveals a 
substantial degradation in overall capabilities. This highlights that
these models have traded general multilingualism for localized expertise. Notably, models curated with DatologyAI achieve competitive results without necessitating such performance tradeoffs. Finally, Figure~\ref{fig:tokens_per_lang} visualizes the performance on
various individual languages as a function of the number of training tokens
in that language
for DatologyAI models and the subset of the models we evaluated where we could obtain reasonable estimates for the per-language training tokens 
(we describe our methodology in Appendix \ref{app:data_efficiency}). This figure clearly visualizes the orders-of-magnitude improvements in 
per-language data efficiency obtained with DatologyAI curation.

Taken together, the results in Figure~\ref{fig:scaling_results} and 
Figures~\ref{fig:per_language_latin}--\ref{fig:tokens_per_lang}
demonstrate that DatologyAI multilingual data curation is both highly effective
and scales to the frontier model training regime. 
The latter point is reinforced by results from Trinity Large, which was pretrained on 17T tokens drawn from the broader DatologyAI-curated corpus and exhibits exceptionally strong multilingual performance.
\section{Conclusion}

Multilinguality is an essential capability for modern foundation models, yet achieving high-quality multilingual performance with broad coverage remains challenging due to uneven data availability across languages and the so-called ``curse of multilinguality".
In this work, we revisit multilingual pretraining from a data-centric perspective and show that many of observed constraints and regressions are not inherent to multilinguality, but instead reflect deficiencies in data quality and curation. 

Through controlled bilingual experiments, we demonstrate that cross-lingual transfer is strongly mediated by data quality: improving English curation alone yields consistent gains across nearly all non-English languages examined (3.91\% average relative improvement across multilingual MMLU, ARC-Challenge, and Belebele), while improving non-English curation reciprocally benefits English performance (1.21\% average relative improvement).
These findings challenge
a purely zero-sum framing of multilingual modeling: higher-quality training data can strengthen multilingual capability without requiring commensurate sacrifices elsewhere in the mixture.

English curation alone, however, is insufficient for optimal performance. Bespoke, per-language pipelines tailored to linguistic and distributional properties deliver substantially larger gains, reaching 16.87\% relative improvement in controlled settings. We further show that translation is most effective when it preserves source quality: translating score-filtered English documents yields materially larger gains than translating arbitrary text, and integrating high-quality document translation as part of a holistic multilingual curation strategy yields far superior results overall.


We productionized these principles through a 20T-token general-purpose pretraining corpus, whose multilingual component was constructed using the curation strategies explored here. Under a controlled 1T-token training budget, 3B and 8B models achieve comparable or stronger multilingual performance than competitive open-weight baselines at 4–10× lower training compute, redefining the multilingual performance–compute Pareto frontier. These efficiency gains persist at frontier scale: Trinity Large Base (400B/A13B), trained on 17T tokens of this corpus, exhibits exceptionally strong multilingual performance relative to its FLOPs budget, validating that the curation principles described here remain effective in the multi–tens-of-trillions regime. We emphasize that for both the 1T-token training budget experiments as well as for Trinity Large, the multilingual performance is obtained using a comparatively minor multilingual token budget of 7.75\% of total training tokens.

Several avenues for future work follow. Our results motivate more systematic, compute-aware mixture design, including per-language sampling strategies and phased curricula that balance improvements in one language against interference in others while ensuring adequate support for low-resource languages. Scaling this agenda will likely require more robust multilingual evaluation frameworks \citep{liang2022helm}. Finally, extending these data-centric principles to multimodal and vision–language model (VLM) training remains an important direction, where evaluation quality and coverage are also central bottlenecks \citep{joshi2026datbench}.

In conclusion, viewed through the data-centric lens advanced in this work, multilinguality need not be a curse of scale, but instead an opportunity to leverage language-aware curation to achieve inclusive, capable foundation models.
\section{Contributions and Acknowledgements}
\label{sec:contri}

\begin{center}
   {\footnotesize\textit{Core and technical contributors listed alphabetically.}}
   \end{center}

\begin{tabularx}{\textwidth}{@{}p{0.25\textwidth}X@{}}
\textbf{Core Contributors} &   Aldo Gael Carranza, Kaleigh Mentzer, and Ricardo Pio Monti \\[0.25em]
\noalign{\vspace{0.75em}}

\textbf{Technical Contributors} & Alex Fang, Alvin Deng, Amro Abbas, Anshuman Suri, Brett Larsen, Cody Blakeney, Darren Teh, David Schwab, Diego Kiner, Fan Pan, Haakon Mongstad, Haoli Yin, Jack Urbanek, Jason Lee, Jason Telanoff, Josh Wills, Luke Merrick, Maximilian Böther, Parth Doshi, Paul Burstein, Pratyush Maini, Rishabh Adiga, Spandan Das, Siddharth Joshi, Tony Jiang, Vineeth Dorna, and Zhengping Wang \\
\noalign{\vspace{0.25em}}
\noalign{\vspace{0.75em}}

\textbf{Leadership} & Bogdan Gaza, Ari Morcos, and Matthew Leavitt \\[0.25em]
\noalign{\vspace{0.75em}}

\textbf{Acknowledgements} & Liz Gatapia,
Jacqueline Liu, Tiffanie Pham, Sylvia Hoang, Kylie Clement, Elise Clark \\
\noalign{\vspace{0.25em}} 
\end{tabularx}

\clearpage
\bibliographystyle{abbrvnat}  
\bibliography{references}

\clearpage
\appendix
\appendix

\clearpage
\section{Appendix}\label{app:main_results}



\subsection{Evaluation datasets per language} \label{app:evals_per_lang}

In the table below, Global MMLU is the dataset provided by \cite{singh2025global} while 
Indic MMLU refers to the translation into Indic languages\footnote{available here: \href{sarvam}{https://huggingface.co/datasets/sarvamai/mmlu-indic}}.
For ARC evaluations, we rely on
evaluation datasets released as part of the Okapi framework \citep{lai2023okapi}. This contains evaluations for the majority of the languages we consider, with the exception of
Korean, Portuguese, Hindi and Bengali. 
For Korean, we use the Ko-ARC evaluation \citep{mcrlkorean2025}, for Portuguese
we use the translated version provided by LumiOpen\footnote{available here: \href{lumi}{https://huggingface.co/datasets/LumiOpen/arc\_challenge\_mt}}. Finally, for Indic ARC evaluations we use Indic ARC\footnote{available here: 
\href{sarvam}{https://huggingface.co/datasets/sarvamai/arc-challenge-indic}}.
In the case of Belebele, the original evaluation dataset supports all our languages \citep{bandarkar2024belebele}\footnote{availabel here: \href{belebele}{https://huggingface.co/datasets/facebook/belebele}}.

\begin{table}[h]
\centering
\begin{tabular}{lllll}
\hline
 \textbf{Language} & \textbf{MMLU} & \textbf{ARC} & \textbf{Belebele} \\
\hline
 Spanish     & Global MMLU & Okapi & Belebele \\
 Portuguese  & Global MMLU & LumiOpen & Belebele \\
 French      & Global MMLU & Okapi & Belebele \\
 German      & Global MMLU & Okapi & Belebele \\
 Italian     & Global MMLU & Okapi & Belebele \\
 Vietnamese  & Global MMLU & Okapi & Belebele \\
 Indonesian  & Global MMLU  & Okapi & Belebele \\
\hline
 Russian     & Global MMLU  & Okapi & Belebele \\
\hline
 Arabic      & Global MMLU  & Okapi & Belebele \\
\hline
 Hindi       & Indic MMLU & Indic ARC & Belebele \\
 Bengali     & Indic MMLU  & Indic ARC & Belebele \\
\hline
 Chinese     & Global MMLU  & Okapi & Belebele \\
 Japanese    & Global MMLU  & Not present & Belebele \\
 Korean      & Global MMLU  & Ko-ARC & Belebele \\
\hline
\end{tabular}
\caption{Table describing the choice of evaluation datasets. }
\end{table}

\clearpage

\subsection{Details of FLOP budget computations for open-source models}
\label{app:flopsflopsflops}

We summarize the training compute (in FLOPs) for each open-source baseline
reported in Figure~\ref{fig:scaling_results}. Throughout this work, we estimate
training FLOPs using the standard approximation
\[
\text{Total FLOPs} \approx 6 \times N \times D,
\]
where \(N\) is the number of (trainable) parameters and \(D\) is the number of
training tokens. In the table, \(\mathrm{B}=10^9\) and \(\mathrm{T}=10^{12}\).
For MoE models, we use the \emph{active} parameter count per token as \(N\).

\begin{table}[h]
\centering
\begin{tabular}{lccl}
\hline
\textbf{Model} & \textbf{Total parameters ($N$)} & \textbf{Total tokens ($D$)} & \textbf{FLOPs ($\approx 6ND$)} \\
\hline
DatologyAI 3B            & 3B  & 1T  & $1.8 \times 10^{22}$ \\
DatologyAI 8B            & 8B  & 1T  & $4.8 \times 10^{22}$ \\
Llama-3.2-1B             & 1B  & 9T  & $5.4 \times 10^{22}$ \\
Llama-3.2-3B             & 3B  & 9T  & $1.6 \times 10^{23}$ \\
Llama-3.2-8B             & 8B  & 15T & $7.2 \times 10^{23}$ \\
SmolLM3-3B               & 3B  & 11T & $1.9 \times 10^{23}$ \\
Granite-4.0-microbase    & 3B  & 15T & $2.7 \times 10^{23}$ \\
Qwen3-1.7B               & 1.7B  & 36T & $3.7 \times 10^{23}$ \\
Qwen3-4B                 & 4B  & 36T & $8.6 \times 10^{23}$ \\
Qwen3-8B                 & 8B  & 36T & $1.7 \times 10^{24}$ \\
LFM-2-2.6B              & 2.6B     & 10T  & $1.5 \times 10^{23}$ \\
LFM-2.5-1.2B              & 1.17B     & 28T  & $1.9 \times 10^{23}$ \\
Trillion Labs Tri-7B              & 7B     & 2T  & $9.3 \times 10^{22}$ \\
Sarvam-1              & 2B     & 4T  & $4.8 \times 10^{22}$ \\ 
Trinity Large            & 13B active (400B total) & 17T & $1.3 \times 10^{24}$ \\
\hline
\end{tabular}
\caption{Estimated training compute (FLOPs) for the open-source baselines in Figure~\ref{fig:scaling_results}, computed as $\approx 6ND$. For MoE models, $N$ denotes the number of \emph{active} parameters per token.}
\end{table}

\clearpage

\subsection{Per language evaluation performance } \label{app:perf_per_language}

\begin{figure}[h!]
    \centering
    \includegraphics[width=\textwidth]{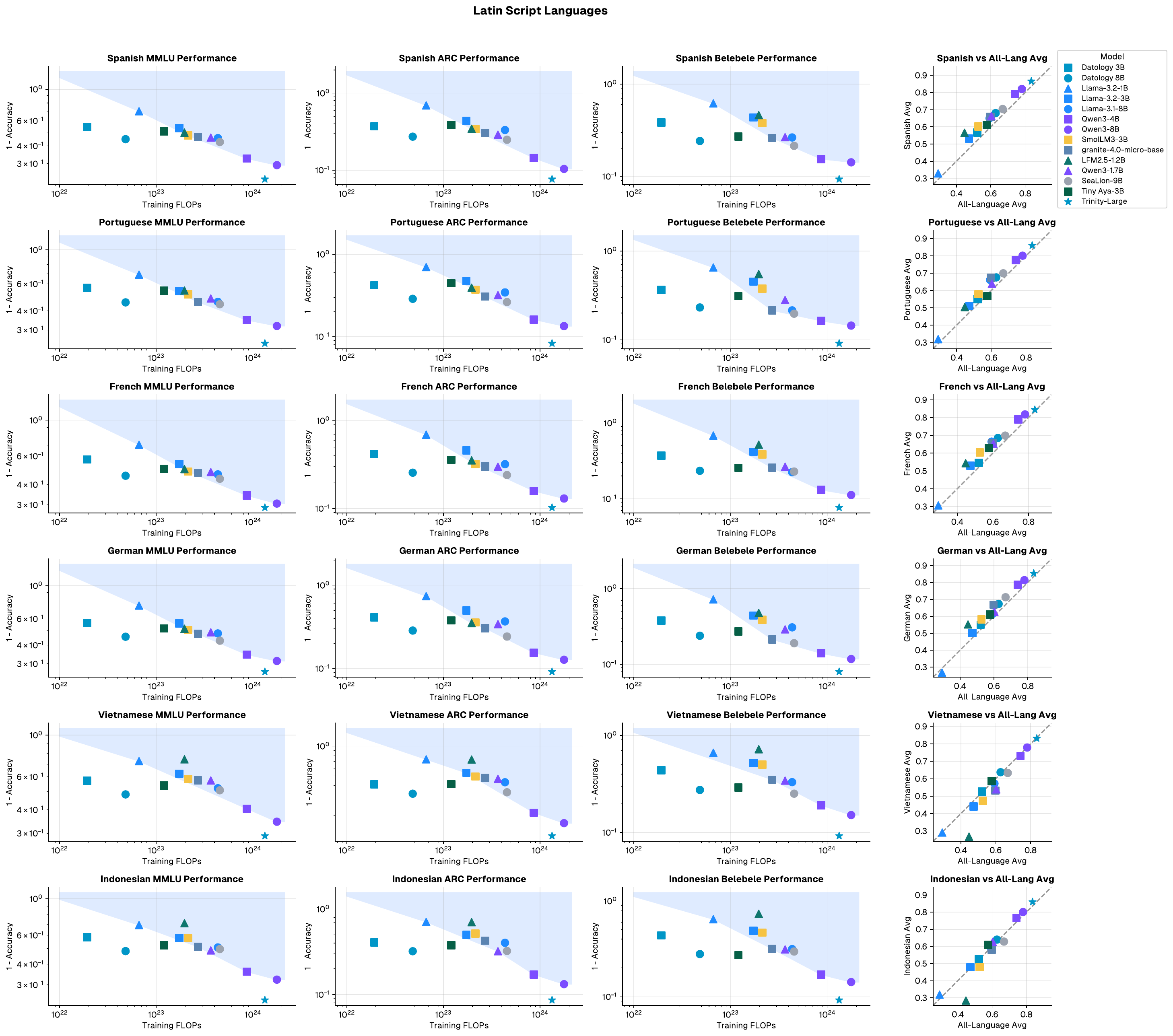}
    \caption{Per-language performance vs. training compute for latin-script languages. Rows correspond to Spanish, Portuguese, French, German, Vietnamese, and Indonesian. Columns 1–3 report performance on MMLU, ARC, and Belebele as a function of training FLOPs (x-axis). The rightmost column compares each model’s language-specific average score (y-axis) to its all-language average across multilingual evaluations (x-axis); the dashed line indicates parity (y = x).}    \label{fig:per_language_latin}
\end{figure}

\clearpage

\begin{figure}[h!]
    \centering
    \includegraphics[width=\textwidth]{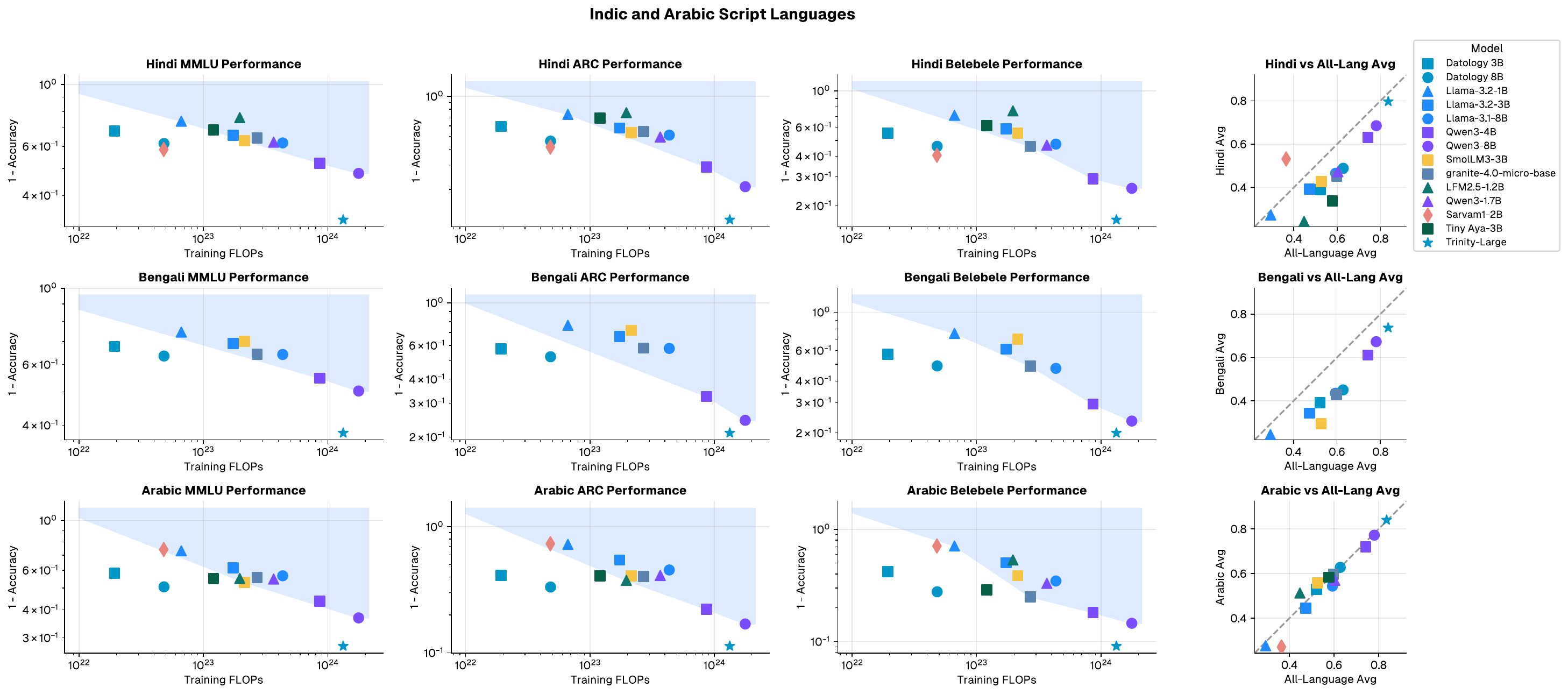}
    \caption{Per-language performance vs. training compute for Indic and Arabic-script languages. Rows correspond to Hindi, Bengali, and Arabic. Columns 1–3 report performance on MMLU, ARC, and Belebele as a function of training FLOPs (x-axis). The rightmost column compares each model’s language-specific average score (y-axis) to its all-language average across multilingual evaluations (x-axis); the dashed line indicates parity (y = x).}    \label{fig:per_language_indic_arabic}
\end{figure}

\begin{figure}[h!]
    \centering
    \includegraphics[width=\textwidth]{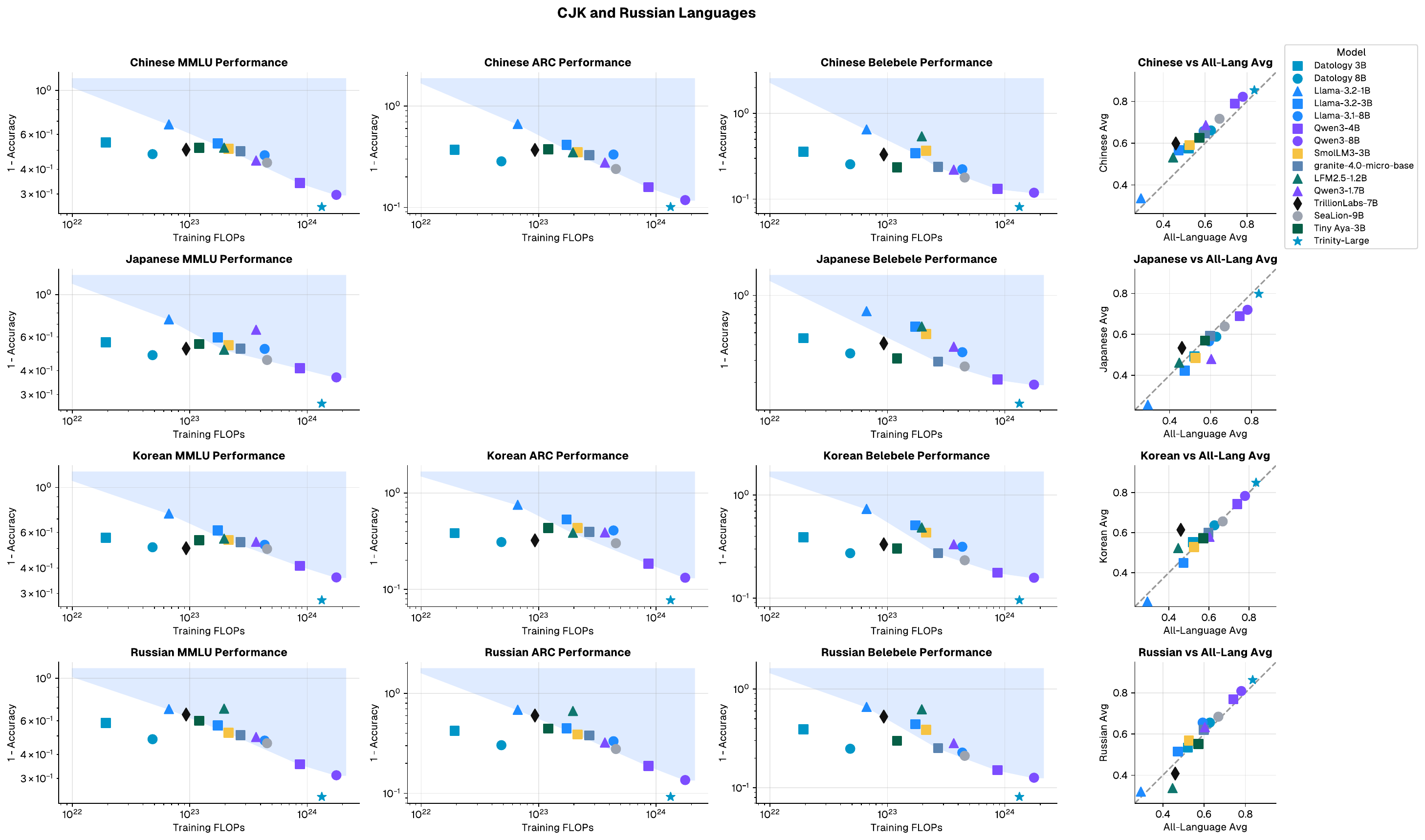}
    \caption{Per-language performance vs. training compute for CJK languages and Russian. Rows correspond to Chinese, Japanese, Korean, and Russian. Columns 1–3 report performance on MMLU, ARC, and Belebele as a function of training FLOPs (x-axis). The rightmost column compares each model’s language-specific average score (y-axis) to its all-language average across multilingual evaluations (x-axis); the dashed line indicates parity (y = x). We note that there is no ARC Challenge evaluation available for Japanese.}    \label{fig:per_language_cjk}
\end{figure}

\clearpage

\subsection{Multilingual data efficiency gains} \label{app:data_efficiency}

In this section, we quantify multilingual performance as a function of the  training token count dedicated to each language. Conducting this analysis is challenging for several reasons. First, several of the evaluated models use their own tokenizers, 
which makes token counts imperfectly comparable across models. 
Second, precise per-language token counts are often 
unavailable for open-source models, and so we aim to estimate them as best we can. 
We nevertheless include this analysis because DatologyAI curation yields improvements in multilingual data efficiency that are large, often by orders of magnitude, 
so the qualitative conclusion is robust even under reasonable uncertainty in these estimates.

In this section we only include models for which we could obtain a reasonably reliable
estimate of per-language tokens using public information. These models are:
\begin{itemize}
    \item SmolLM3: This model used 12\% multilingual data over 11T, supporting a range of languages including Spanish, German, French and Portuguese. We compute the amount of tokens per language directly from the configurations which were publicly shared\footnote{Available at \href{configs}{https://huggingface.co/datasets/HuggingFaceTB/smollm3-configs}}.
    \item Llama3.2: This model used 8\% multilingual data over 9T, supporting seven languages. This is approximately 100B tokens per language.
    \item Sarvam-1: This was was trained on a 2T Indic language corpus, which contained 20\% Hindi tokens and 10\% Bengali tokens. This corresponds to 200B and 100B tokens for each language respectively. 
    \item Trillion Labs 7B: this model was trained on a 2T dataset, 10\% of which was multilingual with a primary focus on Korean. As such, we estimated this model was trained with approximately 200B Korean tokens.
    \item DatologyAI: as referenced in section \ref{section:big_boi_exps}, DatologyAI models were trained for 1T tokens with a 7.75\% multilingual component. This corresponds to a total of 75B multilingual tokens across thirteen languages, approximately 6B tokens per language.  
\end{itemize}

Figure~\ref{fig:tokens_per_lang} visualizes the performance across various languages as a function of estimated tokens in that language. We observe that DatologyAI curation has orders of magnitude gains in token efficiency.

\begin{figure}[h!]
    \centering
    \includegraphics[width=\textwidth]{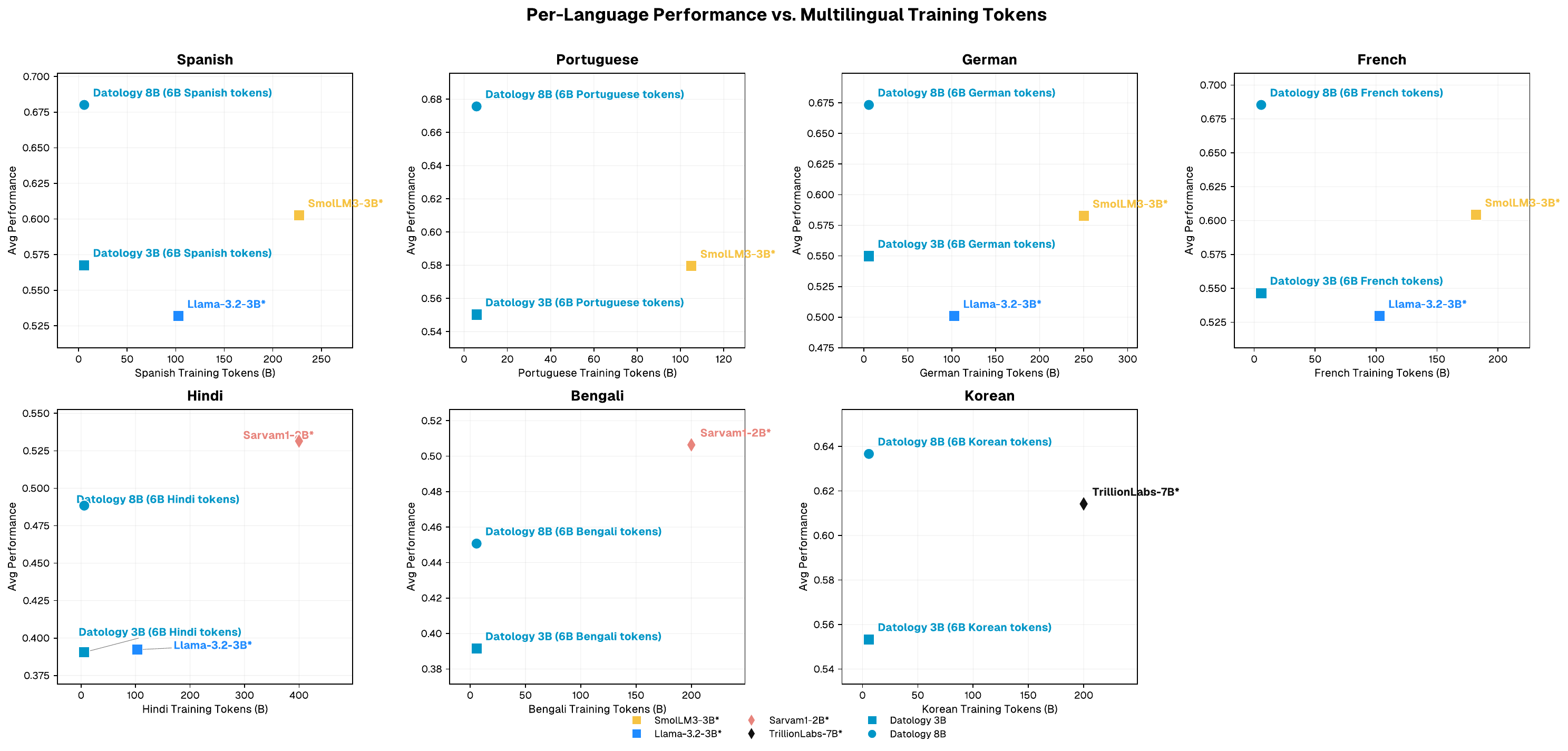}
    \caption{Per-Language Performance vs. Multilingual Training Tokens. 
    We visualize the number of language-specific training tokens (x-axis, billions) and the
    average downstream performance across a range of models. We only include models where the 
    number of tokens per langauge could be reasonably estimated based on public
    information. DatologyAI models were trained with only 6B tokens per language (7.75\% mutlilingual overall).
    The plots demonstrate significant data efficiency improvements from DatologyAI curation compared to 
    open-source baselines such as Llama-3.2, SmolLM2, and language-specific models like Sarvam-2B. We add an asterix beside the name of all non-DatologyAI models, to highlight that we estimated the number of tokens per language to the best of our ability based on publicly available information. 
    }    \label{fig:tokens_per_lang}
\end{figure}

\clearpage

\subsubsection{Numerical results} \label{app:perf_per_language_tables}

Below we report numerical evaluation results por language.

\begin{table}[h]
        \centering
        \begin{tabular}{lccc}
        \hline
        \textbf{Model} & \textbf{MMLU} & \textbf{ARC} & \textbf{Belebele} \\
        \hline
        Datology 3B & 0.46 & 0.63 & 0.62 \\
Datology 8B & 0.55 & 0.73 & 0.76 \\
Llama-3.2 1B & 0.30 & 0.31 & 0.38 \\
Llama-3.2 3B & 0.47 & 0.56 & 0.56 \\
Llama-3.1 8B & 0.55 & 0.67 & 0.74 \\
Qwen3 4B & 0.67 & 0.86 & 0.85 \\
Qwen3 8B & 0.71 & 0.90 & 0.86 \\
SmolLM3 3B & 0.53 & 0.66 & 0.62 \\
Granite-4.0 Micro & 0.54 & 0.70 & 0.74 \\
LFM2.5 1.2B & 0.50 & 0.66 & 0.54 \\
Trinity Large (MoE) & 0.77 & 0.92 & 0.91 \\
        \hline
        \end{tabular}
        \caption{Evaluations for language Spanish}
        \end{table}

\begin{table}[h]
        \centering
        \begin{tabular}{lccc}
        \hline
        \textbf{Model} & \textbf{MMLU} & \textbf{ARC} & \textbf{Belebele} \\
        \hline
        Datology 3B & 0.44 & 0.58 & 0.64 \\
Datology 8B & 0.55 & 0.71 & 0.77 \\
Llama-3.2 1B & 0.31 & 0.30 & 0.35 \\
Llama-3.2 3B & 0.46 & 0.53 & 0.55 \\
Llama-3.1 8B & 0.54 & 0.66 & 0.79 \\
Qwen3 4B & 0.65 & 0.84 & 0.84 \\
Qwen3 8B & 0.68 & 0.87 & 0.86 \\
SmolLM3 3B & 0.49 & 0.63 & 0.62 \\
Granite-4.0 Micro & 0.54 & 0.69 & 0.79 \\
LFM2.5 1.2B & 0.46 & 0.61 & 0.45 \\
Trinity Large (MoE) & 0.75 & 0.92 & 0.91 \\
        \hline
        \end{tabular}
        \caption{Evaluations for language Portuguese}
        \end{table}

\begin{table}[h]
        \centering
        \begin{tabular}{lccc}
        \hline
        \textbf{Model} & \textbf{MMLU} & \textbf{ARC} & \textbf{Belebele} \\
        \hline
        Datology 3B & 0.43 & 0.58 & 0.63 \\
Datology 8B & 0.55 & 0.74 & 0.77 \\
Llama-3.2 1B & 0.29 & 0.31 & 0.32 \\
Llama-3.2 3B & 0.46 & 0.54 & 0.58 \\
Llama-3.1 8B & 0.54 & 0.68 & 0.78 \\
Qwen3 4B & 0.66 & 0.84 & 0.87 \\
Qwen3 8B & 0.69 & 0.87 & 0.89 \\
SmolLM3 3B & 0.52 & 0.68 & 0.61 \\
Granite-4.0 Micro & 0.53 & 0.70 & 0.74 \\
LFM2.5 1.2B & 0.50 & 0.65 & 0.48 \\
Trinity Large (MoE) & 0.71 & 0.90 & 0.92 \\
        \hline
        \end{tabular}
        \caption{Evaluations for language French}
        \end{table}

\begin{table}[h]
        \centering
        \begin{tabular}{lccc}
        \hline
        \textbf{Model} & \textbf{MMLU} & \textbf{ARC} & \textbf{Belebele} \\
        \hline
        Datology 3B & 0.44 & 0.59 & 0.62 \\
Datology 8B & 0.54 & 0.71 & 0.76 \\
Llama-3.2 1B & 0.26 & 0.26 & 0.28 \\
Llama-3.2 3B & 0.44 & 0.50 & 0.56 \\
Llama-3.1 8B & 0.52 & 0.63 & 0.69 \\
Qwen3 4B & 0.65 & 0.85 & 0.86 \\
Qwen3 8B & 0.69 & 0.87 & 0.88 \\
SmolLM3 3B & 0.50 & 0.64 & 0.61 \\
Granite-4.0 Micro & 0.52 & 0.69 & 0.79 \\
LFM2.5 1.2B & 0.48 & 0.65 & 0.52 \\
Trinity Large (MoE) & 0.73 & 0.91 & 0.92 \\
        \hline
        \end{tabular}
        \caption{Evaluations for language German}
        \end{table}

\begin{table}[h]
        \centering
        \begin{tabular}{lccc}
        \hline
        \textbf{Model} & \textbf{MMLU} & \textbf{ARC} & \textbf{Belebele} \\
        \hline
        Datology 3B & 0.43 & 0.59 & 0.56 \\
Datology 8B & 0.52 & 0.67 & 0.73 \\
Llama-3.2 1B & 0.27 & 0.26 & 0.34 \\
Llama-3.2 3B & 0.38 & 0.46 & 0.48 \\
Llama-3.1 8B & 0.48 & 0.57 & 0.67 \\
Qwen3 4B & 0.59 & 0.79 & 0.81 \\
Qwen3 8B & 0.65 & 0.83 & 0.85 \\
SmolLM3 3B & 0.42 & 0.51 & 0.50 \\
Granite-4.0 Micro & 0.43 & 0.52 & 0.65 \\
LFM2.5 1.2B & 0.26 & 0.27 & 0.28 \\
Trinity Large (MoE) & 0.71 & 0.88 & 0.91 \\
        \hline
        \end{tabular}
        \caption{Evaluations for language Vietnamese}
        \end{table}

\begin{table}[h]
        \centering
        \begin{tabular}{lccc}
        \hline
        \textbf{Model} & \textbf{MMLU} & \textbf{ARC} & \textbf{Belebele} \\
        \hline
        Datology 3B & 0.42 & 0.59 & 0.56 \\
Datology 8B & 0.52 & 0.68 & 0.72 \\
Llama-3.2 1B & 0.31 & 0.29 & 0.36 \\
Llama-3.2 3B & 0.42 & 0.50 & 0.51 \\
Llama-3.1 8B & 0.49 & 0.60 & 0.69 \\
Qwen3 4B & 0.64 & 0.83 & 0.83 \\
Qwen3 8B & 0.68 & 0.87 & 0.86 \\
SmolLM3 3B & 0.42 & 0.48 & 0.53 \\
Granite-4.0 Micro & 0.49 & 0.57 & 0.68 \\
LFM2.5 1.2B & 0.29 & 0.30 & 0.26 \\
Trinity Large (MoE) & 0.76 & 0.91 & 0.91 \\
        \hline
        \end{tabular}
        \caption{Evaluations for language Indonesian}
        \end{table}

\begin{table}[h]
        \centering
        \begin{tabular}{lccc}
        \hline
        \textbf{Model} & \textbf{MMLU} & \textbf{ARC} & \textbf{Belebele} \\
        \hline
        Datology 3B & 0.32 & 0.41 & 0.45 \\
Datology 8B & 0.39 & 0.54 & 0.54 \\
Llama-3.2 1B & 0.26 & 0.27 & 0.29 \\
Llama-3.2 3B & 0.34 & 0.42 & 0.41 \\
Llama-3.1 8B & 0.38 & 0.49 & 0.53 \\
Qwen3 4B & 0.48 & 0.71 & 0.71 \\
Qwen3 8B & 0.52 & 0.79 & 0.74 \\
SmolLM3 3B & 0.37 & 0.47 & 0.45 \\
Granite-4.0 Micro & 0.36 & 0.46 & 0.54 \\
LFM2.5 1.2B & 0.24 & 0.25 & 0.24 \\
Trinity Large (MoE) & 0.67 & 0.88 & 0.84 \\
Sarvam1 2B & 0.42 & 0.58 & 0.59 \\
        \hline
        \end{tabular}
        \caption{Evaluations for language Hindi}
        \end{table}

\begin{table}[h]
        \centering
        \begin{tabular}{lccc}
        \hline
        \textbf{Model} & \textbf{MMLU} & \textbf{ARC} & \textbf{Belebele} \\
        \hline
        Datology 3B & 0.32 & 0.42 & 0.43 \\
Datology 8B & 0.36 & 0.48 & 0.51 \\
Llama-3.2 1B & 0.25 & 0.24 & 0.25 \\
Llama-3.2 3B & 0.31 & 0.33 & 0.39 \\
Llama-3.1 8B & 0.36 & 0.42 & 0.53 \\
Qwen3 4B & 0.45 & 0.67 & 0.71 \\
Qwen3 8B & 0.50 & 0.76 & 0.77 \\
SmolLM3 3B & 0.30 & 0.28 & 0.30 \\
Granite-4.0 Micro & 0.36 & 0.42 & 0.51 \\
Trinity Large (MoE) & 0.62 & 0.79 & 0.80 \\
Sarvam1 2B & 0.41 & 0.56 & 0.55 \\
        \hline
        \end{tabular}
        \caption{Evaluations for language Bengali}
        \end{table}

\begin{table}[h]
        \centering
        \begin{tabular}{lccc}
        \hline
        \textbf{Model} & \textbf{MMLU} & \textbf{ARC} & \textbf{Belebele} \\
        \hline
        Datology 3B & 0.42 & 0.59 & 0.58 \\
Datology 8B & 0.49 & 0.67 & 0.72 \\
Llama-3.2 1B & 0.27 & 0.27 & 0.29 \\
Llama-3.2 3B & 0.38 & 0.46 & 0.50 \\
Llama-3.1 8B & 0.43 & 0.55 & 0.65 \\
Qwen3 4B & 0.56 & 0.78 & 0.82 \\
Qwen3 8B & 0.63 & 0.83 & 0.85 \\
SmolLM3 3B & 0.47 & 0.59 & 0.61 \\
Granite-4.0 Micro & 0.44 & 0.60 & 0.75 \\
LFM2.5 1.2B & 0.45 & 0.62 & 0.47 \\
Trinity Large (MoE) & 0.72 & 0.89 & 0.91 \\
        \hline
        \end{tabular}
        \caption{Evaluations for language Arabic}
        \end{table}

\begin{table}[h]
        \centering
        \begin{tabular}{lccc}
        \hline
        \textbf{Model} & \textbf{MMLU} & \textbf{ARC} & \textbf{Belebele} \\
        \hline
        Datology 3B & 0.45 & 0.63 & 0.64 \\
Datology 8B & 0.52 & 0.72 & 0.74 \\
Llama-3.2 1B & 0.33 & 0.33 & 0.35 \\
Llama-3.2 3B & 0.46 & 0.58 & 0.66 \\
Llama-3.1 8B & 0.53 & 0.67 & 0.78 \\
Qwen3 4B & 0.66 & 0.84 & 0.87 \\
Qwen3 8B & 0.70 & 0.88 & 0.88 \\
SmolLM3 3B & 0.49 & 0.65 & 0.63 \\
Granite-4.0 Micro & 0.51 & 0.67 & 0.76 \\
LFM2.5 1.2B & 0.49 & 0.65 & 0.46 \\
Trinity Large (MoE) & 0.74 & 0.90 & 0.92 \\
        \hline
        \end{tabular}
        \caption{Evaluations for language Chinese}
        \end{table}

\begin{table}[h]
        \centering
        \begin{tabular}{lcc}
        \hline
        \textbf{Model} & \textbf{MMLU} & \textbf{Belebele} \\
        \hline
        Datology 3B & 0.44 & 0.55 \\
Datology 8B & 0.52 & 0.66 \\
Llama-3.2 1B & 0.26 & 0.25 \\
Llama-3.2 3B & 0.40 & 0.44 \\
Llama-3.1 8B & 0.48 & 0.65 \\
Qwen3 4B & 0.59 & 0.79 \\
Qwen3 8B & 0.63 & 0.81 \\
SmolLM3 3B & 0.46 & 0.51 \\
Granite-4.0 Micro & 0.48 & 0.71 \\
LFM2.5 1.2B & 0.49 & 0.44 \\
Trinity Large (MoE) & 0.73 & 0.87 \\
        \hline
        \end{tabular}
        \caption{Evaluations for language Japanese}
        \end{table}

\begin{table}[h]
        \centering
        \begin{tabular}{lccc}
        \hline
        \textbf{Model} & \textbf{MMLU} & \textbf{ARC} & \textbf{Belebele} \\
        \hline
        Datology 3B & 0.43 & 0.62 & 0.61 \\
Datology 8B & 0.49 & 0.69 & 0.73 \\
Llama-3.2 1B & 0.26 & 0.25 & 0.26 \\
Llama-3.2 3B & 0.39 & 0.47 & 0.49 \\
Llama-3.1 8B & 0.48 & 0.59 & 0.68 \\
Qwen3 4B & 0.59 & 0.81 & 0.82 \\
Qwen3 8B & 0.64 & 0.87 & 0.84 \\
SmolLM3 3B & 0.45 & 0.57 & 0.57 \\
Granite-4.0 Micro & 0.46 & 0.61 & 0.73 \\
LFM2.5 1.2B & 0.44 & 0.61 & 0.51 \\
Trinity Large (MoE) & 0.72 & 0.92 & 0.90 \\
TrillionLabs 7B & 0.50 & 0.68 & 0.67 \\
        \hline
        \end{tabular}
        \caption{Evaluations for language Korean}
        \end{table}

\begin{table}[h]
        \centering
        \begin{tabular}{lccc}
        \hline
        \textbf{Model} & \textbf{MMLU} & \textbf{ARC} & \textbf{Belebele} \\
        \hline
        Datology 3B & 0.41 & 0.58 & 0.61 \\
Datology 8B & 0.52 & 0.70 & 0.75 \\
Llama-3.2 1B & 0.31 & 0.31 & 0.34 \\
Llama-3.2 3B & 0.43 & 0.55 & 0.56 \\
Llama-3.1 8B & 0.53 & 0.67 & 0.77 \\
Qwen3 4B & 0.64 & 0.81 & 0.85 \\
Qwen3 8B & 0.69 & 0.86 & 0.87 \\
SmolLM3 3B & 0.48 & 0.61 & 0.61 \\
Granite-4.0 Micro & 0.50 & 0.62 & 0.75 \\
LFM2.5 1.2B & 0.31 & 0.33 & 0.38 \\
Trinity Large (MoE) & 0.76 & 0.91 & 0.92 \\
        \hline
        \end{tabular}
        \caption{Evaluations for language Russian}
        \end{table}

\end{document}